\documentclass[sigconf]{acmart}

\copyrightyear{2025}
\acmYear{2025}
\setcopyright{cc}
\setcctype{CC-BY}
\acmConference[SAC '25]{The 40th ACM/SIGAPP Symposium on Applied Computing}{March 31-April 4, 2025}{Catania, Italy}
\acmBooktitle{The 40th ACM/SIGAPP Symposium on Applied Computing (SAC '25), March 31-April 4, 2025, Catania, Italy}\acmDOI{10.1145/3672608.3707764}
\acmISBN{979-8-4007-0629-5/25/03}

\begin{document}
\title{Beyond Text-to-Text: An Overview of Multimodal and Generative Artificial Intelligence for Education Using Topic Modeling}
  
\renewcommand{\shorttitle}{Beyond Text-to-Text}

\author{Ville Heilala}
\orcid{orcid.org/0000-0003-2068-2777}
\affiliation{%
  \institution{University of Jyväskylä}
  \city{Jyväskylä} 
  \country{Finland}
}

\author{Roberto Araya}
\orcid{0000-0003-2598-8994}
\affiliation{%
  \institution{Universidad de Chile}
  \city{Santiago} 
  \country{Chile}
}

\author{Raija Hämäläinen}
\orcid{0000-0002-3248-9619}
\affiliation{%
  \institution{University of Jyväskylä}
  \city{Jyväskylä} 
  \country{Finland}
}

\renewcommand{\shortauthors}{Heilala et al.}

\begin{abstract}

Generative artificial intelligence (GenAI) can reshape education and learning. While large language models (LLMs) like ChatGPT dominate current educational research, multimodal capabilities—such as text-to-speech and text-to-image—are less explored. This study uses topic modeling to map the research landscape of multimodal and generative AI in education. An extensive literature search yielded 4175 articles. Employing a topic modeling approach, latent topics were extracted, resulting in 38 interpretable topics organized into 14 thematic areas. Findings indicate a predominant focus on text-to-text models in educational contexts, with other modalities underexplored, overlooking the broader potential of multimodal approaches. The results suggest a research gap, stressing the importance of more balanced attention across different AI modalities and educational levels. In summary, this research provides an overview of current trends in generative AI for education, underlining opportunities for future exploration of multimodal technologies to fully realize the transformative potential of artificial intelligence in education.

\end{abstract}

%
%



\maketitle

\section{Introduction}

Whether you have experience in Education 2.0 \citep{Cope-and-Mary-Kalantzis2019-sx}, Education 3.0 \citep{Vrabie2023-oi}, Education 4.0 \citep{Pelaez-Sanchez2024-qt}, Education 5.0 \citep{Matthew2024-ti}, or even already in Education 6.0 \citep{Moleka2023-ri}, you might have thought of how artificial intelligence (AI) could acceptably and responsibly \citep{Nemorin2023-jq} realize its ``potential for reshaping the core foundations of education, teaching and learning'' \citep[][p.~3]{UNESCO2019-nf}. Recently, the UNESCO \textit{Guidance for generative AI in education and research} noted that developments in generative AI might trigger a transformative change in established educational systems and their foundations \citep{UNESCO2023-hu}. Generative AI refers to models capable of producing text, images, video, computer code, and other content modalities by utilizing the data on which they were trained. These advanced technologies applied in educational contexts create opportunities and challenges, with regulatory efforts trying to address emergent issues. At the same time, the uneven distribution of research-based knowledge and the influence of market leaders pose questions for the future of pedagogy. Thus, this study seeks to shed light on artificial intelligence for education by synthesizing research literature using topic modeling to answer the following research question: What is the high-level research landscape of multimodal approaches and generative AI in education?

\section{Background}

Various AI-driven pedagogical and technological solutions are emerging at an accelerating pace, creating both possibilities and tensions \citep{Ahmad2024-og,Nemorin2023-jq}. At the same time, organizations and legislators, such as the EU, strive to keep up with opportunities and challenges. For example, the recently published EU Regulation No. 2024/1689 (Artificial Intelligence Act), which aims to harmonize rules on AI, regulates issues such as emotion recognition in educational contexts. From the educational standpoint, a key challenge arises when research-based knowledge is unevenly distributed across different perspectives and technologies, with market leaders heavily impacting the landscape of AI in education \citep[e.g.,][]{Verdegem2022-an}. For instance, if technological advances primarily drive various pedagogical approaches, it is crucial to recognize the influence this has. It is equally important to discern what aspects of these technologies are hype and which areas risk being overlooked \citep[e.g.,][]{Humble2022-eb,Tan2023-pj,Matre2024-rk,Yusuf2024-io}.

From the historical perspective, the fields of artificial intelligence and education have been closely linked since the inception of AI, with early AI pioneers contributing to both areas by using AI to understand and improve human learning, a perspective that has since been argued to be diminished but still revivable \citep{Doroudi2022-fr}. Also, computational methods for performing transformations between different data modalities have existed, for example, for text-to-text, text-to-speech, and speech-to-text since the 1950s \citep{Hutchins2004-nr,Klatt1987-er,Davis1952-yg} and text to image since the 2000s \citep{Agnese2020-gt}. Recently, deep learning has allowed the development of other transformations like text-to-video, video-to-text \citep{Perez-Martin2022-tt}, and text-to-music \citep{Chen2024-dl}. These multimodal capabilities are suggested to benefit education \citep{Chen2024-vi}, and multimodal AI is claimed to possess even an ``immense'' \citep[][p.~2]{Lu2024-es} and ``vast'' \citep[][p.~271]{Mohamed2024-so} potential. Thus, both AI research and educational research together \citep{Chiu2023-nt} should investigate the multimodal and generative capabilities of these so-called foundation models \citep{Bommasani2021-qt} to ensure ``human-centered and pedagogically appropriate interaction'' \citep[][p.~29]{UNESCO2023-hu}.

Recently, reviews have identified main application areas of AI in education like assessment, administration, prediction, AI assistants, content delivery, intelligent tutoring systems, and managing student learning, while also highlighting challenges such as negative perceptions, technology skills gaps, ethical concerns, and AI tool design issues \citep{Crompton2023-kl,Crompton2024-rn}. Higher education and the educational domains of medical, language, engineering, and computing education have received the most attention in research \citep{Yusuf2024-io,Crompton2023-kl}. Future research should examine transparency, bias mitigation, curriculum design, and ethical usage to enhance educational practices across various disciplines \citep{Bahroun2023-hj}. In general, an extensive body of research on AI in education highlights a broad range of applications, theories, and methodologies \citep[e.g.,][]{Wang2024-zr}.

However, the development of AI technologies for education is still argued to lag behind, with most tools being simple, single-purpose, and not tailored to interdisciplinary learning needs, highlighting the need for researchers to examine and design more advanced, versatile, and useful AI tools for education \citep{Chiu2023-nt,Setala2024-ot}. For example, while the long-existed text-to-speech technology has shown promising potential for improving writing performance, spelling, and reading comprehension among adolescents with learning difficulties, the amount of high-quality research in this area remains limited, indicating the need for further investigation \citep{Matre2024-rk}. In general, multimodal capabilities of generative AI in educational research have received limited attention because existing research has focused primarily on text-to-text models \citep{Chen2024-vi,Matre2024-rk}. Also, recent research \citep[e.g.,][]{Yusuf2024-io,Humble2022-eb} observed that prior studies have often suffered from inconsistencies and weaknesses in the search strategies and ambiguity in the concept of AI. Thus, the novelty of this research was the extensive search string incorporating a relatively broad set of multimodal capabilities of generative AI.

\section{Materials and methods}

\begin{figure}[htbp!]
    \centering
    \includegraphics[width=0.9\linewidth]{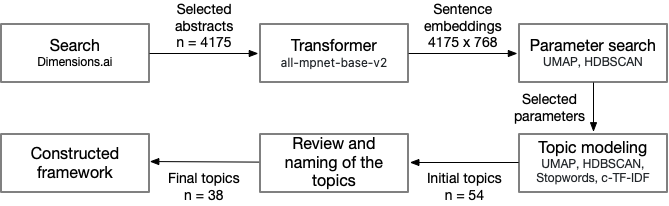}
    \caption{Topic modeling process}
    \label{fig:process}
    \Description[Figure 1]{Topic modeling process}
\end{figure}

\subsection{Data collection}

Data collection was conducted in August 2024 using Dimensions.ai service, and a complementary search was conducted using Scopus and Web of Science (WOS). However, the complementary search did not yield any new results. Dimensions.ai is argued to provide an extensive literature coverage \citep[e.g.,][]{Thelwall2018-dc,Verma2023-tb}. Also, when selecting the example articles, we found that at least one article, \citep{Riabko2024-bi}, was found only using Dimensions.ai.

Focusing on student perspective, the following search string was used to search articles, proceedings, and book chapters based on title and abstract with no limitations to publishing year:

\begin{quote}
\textit{(education OR student OR school) AND ('generative ai' OR 'generative artificial intelligence' OR genai OR LLM OR 'large language model' OR 'text-to-text' OR 'text-to-video' OR 'text-to-audio' <COMBINATIONS>}
\end{quote}

\noindent
The last part of the search string \textit{<COMBINATIONS>} consisted of all combinations of different modalities between text, video, audio, voice, sound, music, data, image, speech, graph, table, emotion, language, and gesture. Notably, to allow more unbiased review of technologies, no service names (e.g., ChatGPT) were included in the search string. The initial search yielded 4475 results. After removing results that did not contain a Document Object Identifier (DOI) or abstract, the final dataset consisted of 4175 articles, forming the text corpus for topic modeling (Figure \ref{fig:process}). 95\% of the articles were published after 2014. The earliest relevant article in this corpus was from 1983, mentioning a computational approach for transforming data modalities in an educational context that dealt with text-to-speech support for visually impaired students \citep{Omotayo1983-zp}.

\subsection{Topic modeling}

To extract latent topics from the text corpus (Figure \ref{fig:process}), we applied the BERTopic approach \citep{Grootendorst2022-ft}, which has been suggested to be able to generate novel insight and encode contextual information by relying on sentence transformers and embeddings \citep{Egger2022-zr}. BERTopic approach has been utilized, for example, to explore dynamics of AI applications \citep{Raman2024-al} and interdisciplinary topics in science \citep{Wang2023-gy}.

Firstly, we created transformer-based embeddings of the abstracts using the general-purpose sentence transformer \textit{all-mpnet-base-v2}\footnote{https://web.archive.org/web/20240823053754/https://huggingface.co/sentence-transformers/all-mpnet-base-v2}. The transformer has been trained, among many other datasets, using over 116 million English-language academic papers from The Semantic Scholar Open Research Corpus (S2ORC)\footnote{https://github.com/allenai/s2orc} \citep{lo-wang-2020-s2orc}. Secondly, the created 768-dimensional document embeddings were then dimensional reduced using Uniform Manifold Approximation and Projection (UMAP) \citep{McInnes2018-zg}. Thirdly, using the reduced embeddings, the documents were clustered by applying the Hierarchical Density-Based Spatial Clustering of Applications with Noise (HDBSCAN) algorithm \citep{Campello2013-sx}. Finally, English stopwords were removed, and the class-based Term Frequency-Inverse Document Frequency (c-TF-IDF) approach was used to generate initial topics. Topics were reviewed and interpreted by the authors.

Both UMAP and HDBSCAN require several hyperparameters to perform dimensionality reduction and clustering. Instead of arbitrarily choosing the parameters, topic count, or topic size, we employed a parameter search approach by randomly sampling combinations of parameters for UMAP and HDBSCAN (Figure \ref{fig:process}). This allowed us to explore various parameters without exhaustively testing every combination. Specifically, we performed 5000 iterations, each time selecting random values for key parameters, such as the number of neighbors (range [2, 55]), minimum distance (0.0 or 0.1), and distance metrics (Manhattan, Euclidean, cosine) for UMAP, as well as the minimum cluster size (range [2, 55]), minimum samples (range [2, 55]), clustering method (leaf, eom), and distance metrics (Manhattan, Euclidean) for HDBSCAN (Figure \ref{fig:params}).

The optimal hyperparameter solution for this research was selected based on the Density-based Clustering Validation (DBCV) index\citep{Moulavi2014-tv}, topic count, and out-of-topic abstract count. The DBCV index is a metric that can be used to assess the quality of clustering by evaluating the density-based separation between clusters and the cohesion within clusters. The optimal solution aimed to obtain a feasible number of topics while leaving the minimum number of abstracts not assigned to any topic. In other words, a clustering solution (i.e., number of topics) for each solution in the set of all clustering solutions \( S \) that minimizes the number of out-of-topic documents and maximizes the DBCV index was assessed as follows:
\[
S^* = \arg\max_{S_i \in S'} D(S_i) \quad \text{where} \quad S' = \arg\min_{S_i \in S} O(S_i),
\] where \( O(S_i) \) represent the number of out-of-topic (i.e., not assigned abstracts) documents for solution \( S_i \) and \( D(S_i) \) represent the DBCV index for solution \( S_i \). Figure \ref{fig:scatter} depicts the solutions in \( S^* \) for solutions having the number of topics ]1, 150[. In this research, the number of topics less than 20 were considered too general, and more than 150 were considered not feasible for a human to interpret qualitatively. In the selected topic model, the interpretability of a single topic was assessed using information entropy: the lower the entropy of a topic, the more likely it is to be considered a valid topic \citep{Wang2023-gy}. Lastly, the topics were arranged as a thematic map combining similar topics to higher-level representations \citep{Braun2006-zy}.

\section{Results}

\subsection{Different modalities}

OpenAI ChatGPT showed the most occurrences in the corpus compared to similar types of generative AI services like Google Gemini or Bard, Microsoft Copilot, Anthropic Claude, and Perplexity.ai (Table \ref{tab:keywords}). However, some keywords can refer to a form of generative AI with various meanings in the literature. The term \textit{copilot} is used, for example, to denote various interactive companions ``that can follow diverse instructions and coherently and accurately answer complex open-ended questions in natural language'' \citep[][p.~2]{Lu2024-es}. On the other hand, copilot can also refer to a specific software like Microsoft Copilot \citep[e.g.,][]{Ghimire2024-iu} or GitHub Copilot \citep[e.g.,][]{Venkatesh2023-gm}. Similarly, the term perplexity can refer to an information theoretical measure \citep[e.g.,][]{Xu2024-ex} or a software called Perplexity.ai \citep[e.g.,][]{Sadeq2024-ir}. While there is a lack of precise conventions for naming used technologies, OpenAI ChatGPT seems to dominate the research concerning LLMs in education, which is not surprising considering the simplicity of its operation \citep[e.g.,][]{Sedaghat2023-di} and prior similar review results \citep[e.g.,][]{Bahroun2023-hj}. Including specific product names in the literature, search string can yield different results, but this research addressed literature that explicitly mentioned different generative models instead of specific software services.

\begin{table}[htpb!]
    \centering
    \caption{The number of titles/abstracts mentioning selected keywords in the corpus (including different forms like llm, llms, generative ai, genai etc.)}
    \begin{tabular}{lc}
        \toprule
        Keyword & Occurrences \\
        \hline
        large language model & 3645 \\
        generative artificial intelligence & 2177 \\
        chatgpt & 1471 \\
        text to speech & 461 \\
        (Continuing in Figure \ref{fig:model-count}) & \\
        \bottomrule
    \end{tabular}
    \label{tab:keywords}
\end{table}

\begin{figure}[htbp!]
    \centering
    \includegraphics[width=0.9\linewidth]{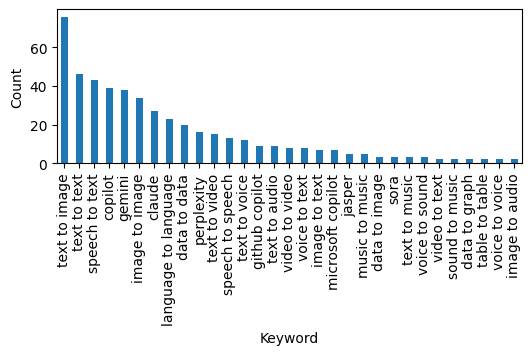}
    \caption{The number ($n \geq 2$) of titles/abstracts mentioning different keywords in the corpus}
    \label{fig:model-count}
    \Description[Figure 2]{The number ($n \geq 2$) of titles/abstracts mentioning different keywords in the corpus}
\end{figure}

Text-to-speech was the next common keyword after language models followed by text-to-image, text-to-text, and speech-to-text, reflecting the maturity of technologies \citep[i.e.,][]{Hutchins2004-nr,Klatt1987-er,Davis1952-yg,Agnese2020-gt} and diffusion of innovations into educational domain (Table \ref{tab:keywords}, Figure \ref{fig:model-count}). Research suggests that text-to-speech and speech-to-text technologies can be used, for example, to support reading comprehension \citep{Sulaimon2023-us} and children who have writing difficulties \citep{Kambouri2023-fn}. Text-to-image technology has been utilized in education, for example, for supporting creative ideation \citep{Liu2024-bv}, facilitating understanding of concepts in structural engineering \citep{Chacon2023-we}, and creating facial images for medical education \citep{Fan2023-iv}. Image-to-image technology has been utilized for creating photorealistic macromolecular visualization for educational purposes \citep{Durrant2022-zd}. The emerging text-to-video and its manifestations like OpenAI Sora were explored for a possibility to augment and personalize learning and facilitate AI-driven content creation \citep{Adetayo2024-nt,Mohamed2024-so}. One multimodal approach was found that used several multimodal approaches to create teaching materials \citep{Chen2024-vi}.

\subsection{Topic modeling}

\begin{figure}[htbp!]
    \centering
    \includegraphics[width=0.89\linewidth]{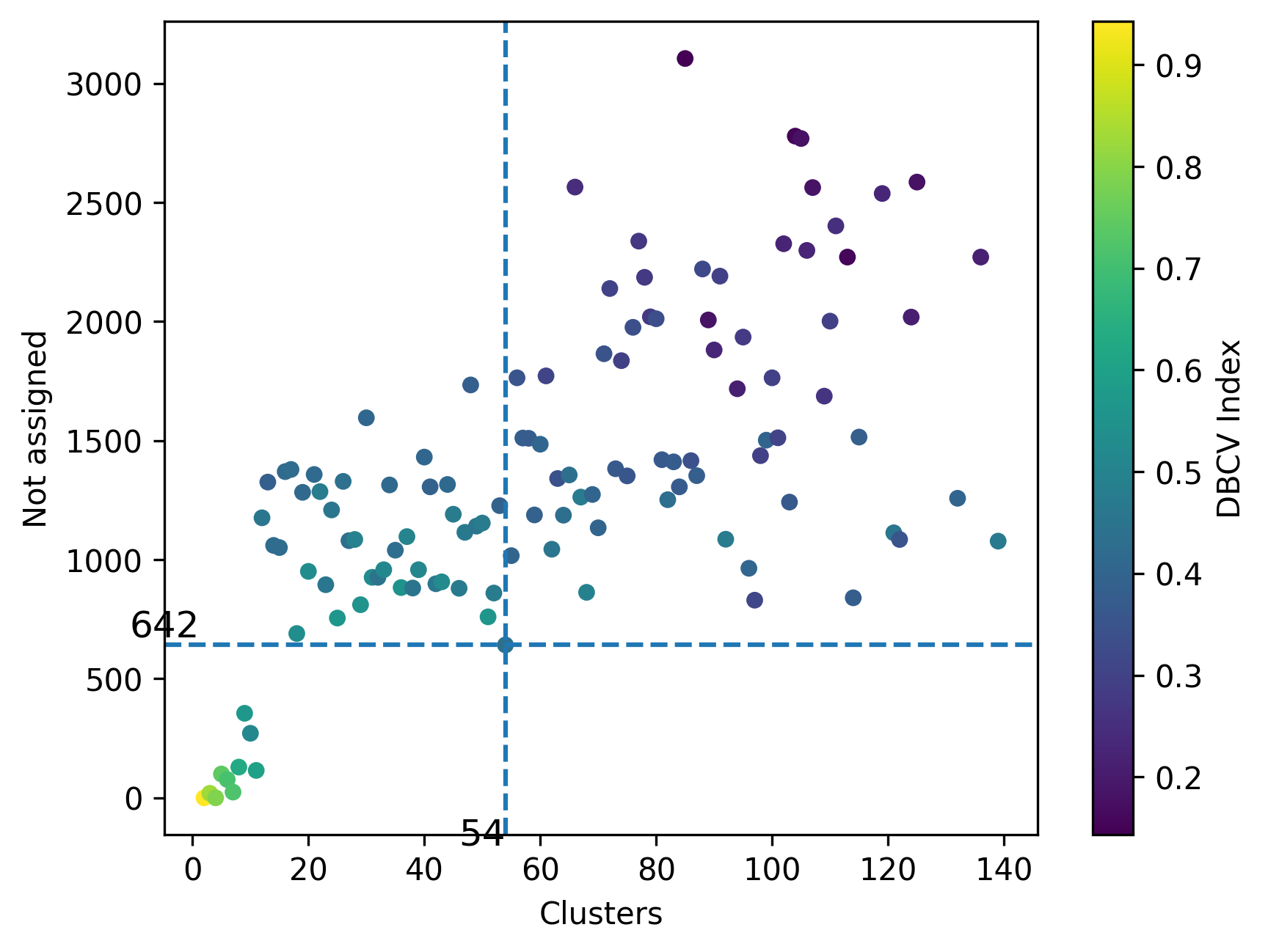}
    \caption{Best topic modeling solution for the number of clusters between ]1, 150[}
    \label{fig:scatter}
    \Description[Figure 3]{Best topic modeling solution for the number of clusters between ]1, 150[}
\end{figure}

\begin{figure*}[htbp!]
    \centering
    \includegraphics[width=0.89\linewidth]{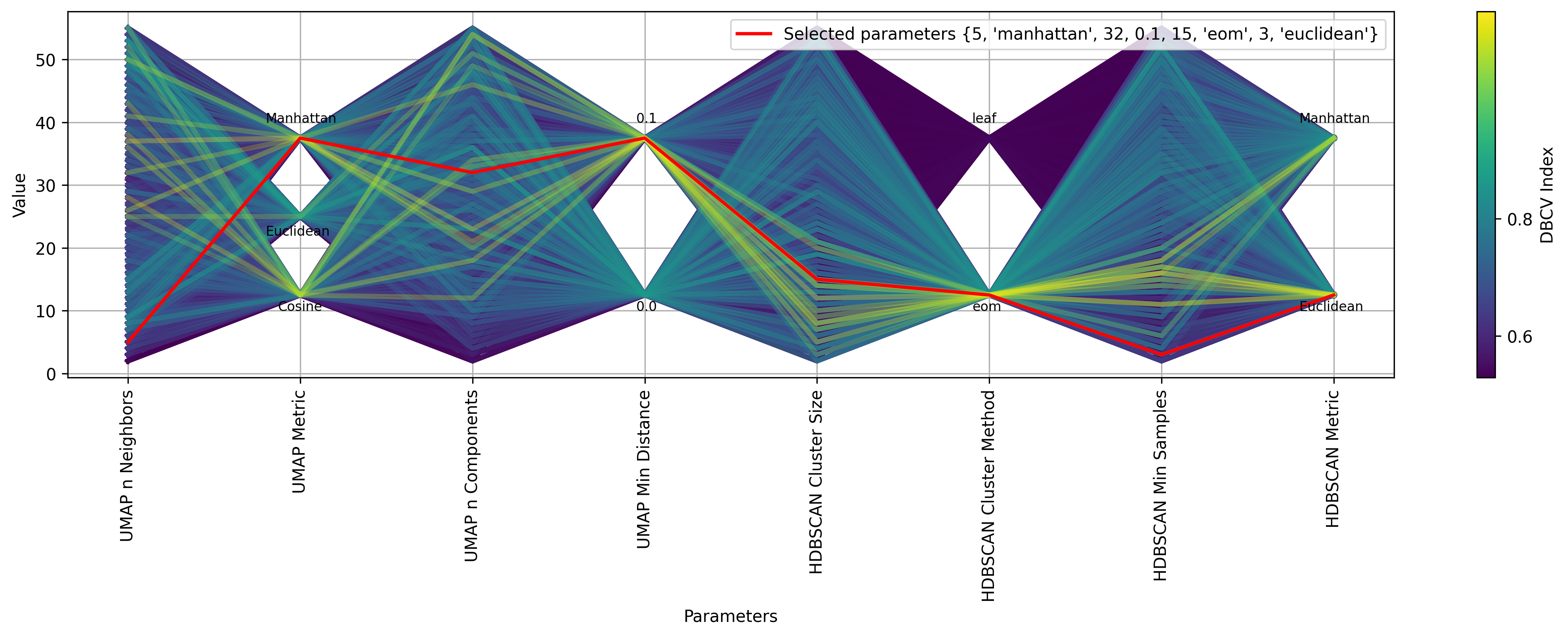}
    \caption{A parallel coordinate plot depicting the hyperparameter space (n=5000) and the selected parameters. One line represents one set of parameters.}
    \label{fig:params}
    \Description[Figure 4]{A parallel coordinate plot depicting the hyperparameter space (n=5000) and the selected parameters. One line represents one set of parameters}
\end{figure*}

A topic modeling approach was used to further summarize the corpus of abstracts. The first step was to obtain a reasonable set of hyperparameters for modeling by randomly sampling combinations of parameters. Figure \ref{fig:scatter} shows the best solutions for each number of clusters (i.e., the number of topics) based on minimizing the number of not assigned abstracts and maximizing the DBCV index. A feasible number of topics, with a topic count of 54, was selected as the optimal solution for obtained hyperparameters and study context. A parallel coordinate plot, Figure \ref{fig:params}, depicts the hyperparameter space showing all the parameter solutions by DBCV index and the selected hyperparameter solution.

Using the selected parameters, topic modeling yielded 54 initial topics consisting of 85\% of the initial corpus of abstracts. Abstracts not assigned to any topic (15\%) were not forced to any topic after modeling because forcing them to the topic model might have introduced additional errors. The information entropy of the topics ranged from 3.083 to 3.367, and none showed significantly higher (2 x SD) information entropy (M = 3.290, SD = 0.062), suggesting that all were equally valid topics. After reviewing and interpreting the topics, eight were merged with others because of similar representation and content. Eight topics were deemed uninterpretable and removed from the final set. Table \ref{tab:topics1} shows the final set of representations and example articles of 38 topics consisting of 74\% of the initial corpus of abstracts. When possible, the example articles were chosen based on the publication year 2024 and open-access publication type, allowing interested readers unrestricted access.

\begin{figure*}[htbp!]
    \centering
    \includegraphics[width=0.93\linewidth]{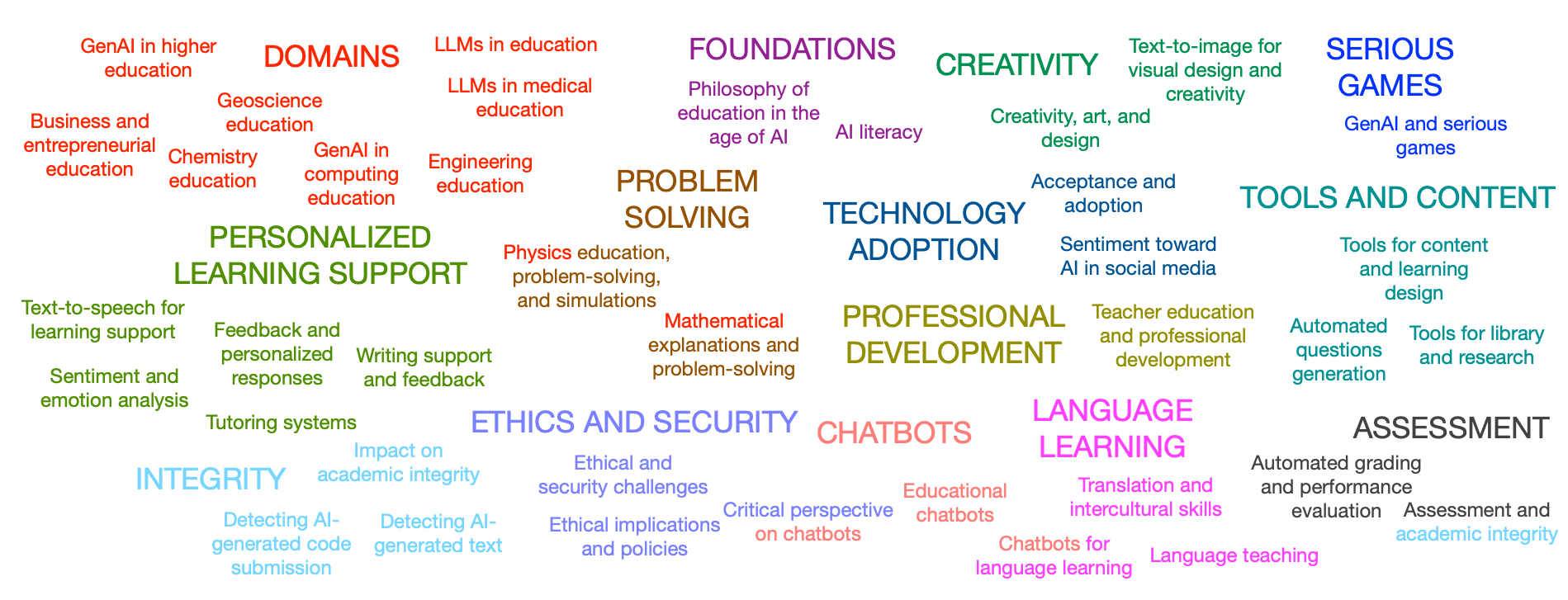}
    \caption{A thematic map of research on generative AI in education based on the topic modeling results}
    \label{fig:topic-map}
    \Description[Figure 5]{A thematic map of research on generative AI in education based on the topic modeling results}
\end{figure*}

The topics were further arranged as a thematic map of research on generative AI in education, depicting 14 prominent thematic focus areas, each with corresponding lower-level topics (Figure \ref{fig:topic-map}). The number of abstracts assigned to each topic is indicative and serves as a general representation, but not all abstracts may perfectly align with or be definitive representatives of the corresponding topic. However, the number of topics identified can provide insights into each research area's relative magnitude and prominence.

Within the \textit{Domains} theme, generative AI was applied across educational domains, for example, to enhance problem-solving in specialized fields like geotechnical engineering \citep[e.g.,][]{Chen2024-ip}, improve learning strategies in chemistry \citep[e.g.,][]{Tassoti2024-dg} and business research education \citep[e.g.,][]{Aure2024-vu}, adapt curriculum design and content creation in engineering education \citep[e.g.,][]{Yelamarthi2024-ru}, support interdisciplinary and AI literacy-focused higher education \citep[e.g.,][]{Chiu2024-ju}, and simplify complex health education materials for improved accessibility \citep[e.g.,][]{Rouhi2024-ku}. The \textit{Personalized Learning Support} theme is related to integrating tools like text-to-speech \citep[e.g.,][]{Weerakoon2024-rg}, sentiment and emotion analysis \citep[e.g.,][]{Parker2024-wr}, and feedback and tutoring systems \citep[e.g.,][]{Liu2024-ua,Lin2024-vf} to enhance individual learning experiences. \textit{Problem Solving} includes areas like mathematical and physics problem-solving \citep[e.g.,][]{Riabko2024-bi}, simulations \citep[e.g.,][]{Farhana2024-lw}, and explanations \citep[e.g.,][]{Collins2024-xg}.

\textit{Technology Adoption} concerns factors influencing the acceptance of AI \citep[e.g.,][]{Shahzad2024-gw}, for example, using sentiment analysis in social media \citep[e.g.,][]{Mamo2024-wp}, while \textit{Professional Development} centers on teacher education and professional preparedness \citep[e.g.,][]{Tan2024-er}. The theme of \textit{Creativity} explores generative AI for facilitating creativity, for example, in visual design and art using technologies like text-to-image generation \citep[e.g.,][]{Vartiainen2023-yt,Jung2024-hg}. \textit{Serious Games} focuses on using generative AI in developing educational games \citep[e.g.,][]{Tyni2024-wd}, and \textit{Tools and Content} addresses the utilization of AI for content design \citep[e.g.,][]{Powell2024-ot}, automated question generation \citep[e.g.,][]{Hang2024-as}, and managing information \citep[e.g.,][]{Deschenes2024-em}.

\textit{Assessment} explores grading, performance evaluation, and assessment using generative AI \citep[e.g.,][]{Gao2024-aw} and the implications for academic integrity \citep[e.g.,][]{Luo-Jess-2024-vi}. \textit{Ethics and security} encompasses concerns on generative AI \citep[e.g.,][]{Zuber2024-ml} and critical perspectives on using AI tools like chatbots \citep[e.g.,][]{Jensen2024-pm} in educational and research contexts, addressing their impact on integrity, privacy, and academic principles \citep[e.g.,][]{Nartey2024-fm}. Closely related \textit{Integrity} theme highlights issues about AI-generated content and its impact on academic integrity \citep[e.g.,][]{Lee2024-qv}, particularly in detecting AI-generated text \citep[e.g.,][]{Fleckenstein2024-ts} and code submissions \citep[e.g.,][]{Pan2024-ys}. The theme \textit{Chatbots} examines the use of AI-driven chatbots for education \citep[e.g.,][]{Hedderich2024-mz,Sikstrom2022-jz}, for example, in language learning \citep[e.g.,][]{Zhang2024-wo}. Lastly, \textit{Language Learning} concerns using AI for language teaching \citep[e.g.,][]{Moorhouse2024-hm}, translation and intercultural skills, and educational interactions \citep[e.g.,][]{Dai2024-ri}.

\section{Discussion and conclusion}

This study aimed to explore the landscape of multimodal and generative AI in education by employing a topic modeling approach, identifying 38 research topics. The thematic analysis synthesized the topics into 14 key focus areas (capitalized and in different colors in Figure \ref{fig:topic-map}), each encompassing various lower-level topics (in lowercase in Figure \ref{fig:topic-map}), offering insights into the prominence and scale of different research domains. The findings suggest that generative AI technologies, particularly OpenAI ChatGPT, dominate educational research, reflecting the focused interest and rapid adoption of LLMs in academic contexts. However, the results also highlight the potential of AI incorporating different modalities (e.g., text-to-speech \citep{Sulaimon2023-us}, speech-to-text \citep{Levine2023-tp}, and text-to-image \citep{Jung2024-hg}) in personalized learning, problem-solving, and creativity, areas that have received less attention compared to LLMs dealing with text-to-text transformations.

The results align and support the findings from the other recent literature, arguing that there is a need to extend the research to other technologies across educational levels. To emphasize, the next common technology after LLMs found in this research, text-to-speech, is argued to have a limited number of high-quality research \citep{Matre2024-rk}. Also, the only educational level identified at a topic level in this research was higher education, which supports the claims that there is limited research on other levels (e.g., K-12) \citep{Yusuf2024-io,Vartiainen2023-yt,Crompton2024-rn}.

Generative AI can challenge the foundational aspects of learning and education. Some authors argue that generative AI reflects a surface approach to learning and a ``functionalist and economist vision of society'' \citep[][p.~231]{Class2024-nj}. Thus, to protect and enhance student agency \citep{Heilala2023-lv,Heilala2022-xt}, we need to have a broad understanding of different AI technologies \citep{UNESCO2023-hu}. For example, while LLMs can offer assistance in drafting and editing \citep{Liu2024-ua}, they also raise questions about academic integrity, as the ease of generating text may compromise the originality of student work \citep{Fleckenstein2024-ts,Luo-Jess-2024-vi}. In other words, there seems to be a tension between using LLMs as supportive tools to help students improve their skills and the growing concerns about plagiarism and the authenticity of student work generated through these technologies. 


For students creating new knowledge in the AI era, the challenge is at the same time mastering the subject matter and reusing knowledge acquired through generative AI \citep{Class2024-nj}. AI for education could greatly benefit from AI models deliberately trained to serve educational needs (e.g., EdGPTs) \citep{UNESCO2023-hu}. Also, integrating multimodal AI-driven tools for visual and auditory content could extend possibilities beyond the capabilities of traditional LLMs by offering richer and more dynamic learning experiences. In addition to medical education, language education, and the natural sciences, fields such as the arts, social sciences, and humanities could also benefit from multimodal approaches in generative AI by enhancing creativity \citep{Vartiainen2023-yt}, critical and creative thinking \citep{Jung2024-hg,Tapia2024-rk}, and interactive learning experiences \citep{Aure2024-vu}. Thus, future research could examine the potential of educationally focused multimodal foundation models that balance knowledge building and knowledge reproduction in different domains of learning end education. Finally, we recommended that educators open-mindedly play around with a broad range of multimodal AI technologies, such as text-to-speech, speech-to-text, and text-to-image. Bravely experimenting with diverse AI technologies could help educators develop their pedagogical AI imagination, enabling them to creatively reimagine pedagogical approaches and design innovative, technology-enhanced learning experiences to enhance personalized learning and support diverse student needs.

This study has its limitations. Topic modeling of a large corpus offers a general and high-level view of the underlying content, leaving more nuanced and emerging topics hidden. The extensive search string used in this study did not capture all variations of product names associated with generative AI technologies. Thus, it is recommended that future research adopt a harmonized approach to describing the underlying technologies to facilitate more comprehensive coverage and clarity in future reviews. Teachers and teaching were not included as keywords in the initial search, and future studies could concentrate more on generative AI from educators' perspectives. Also, future studies could focus on different themes identified in this research, which might provide a more fine-grained view of generative AI in education. The search string included several technologies, but future studies could include more (e.g., text to sign language \citep{Kahlon2023-qt}).

In conclusion, this research provided an overview of the current research landscape on generative AI in education, highlighting the need to extend the scholarly horizon beyond text-to-text technologies. While LLMs dominate the research, other promising technologies like text-to-speech remain underexplored, suggesting the need for more balanced attention across AI modalities. Nevertheless, LLMs have proven to be powerful tools for building knowledge---after all, also this research extensively utilized them.

\begin{table*}[htbp!]
  \caption{Topics, representations, and example articles}
  \label{tab:topics1}
  \begin{tabular}{p{0.2cm} p{0.4cm} p{3cm} p{11.3cm} p{0.9cm}}
    \toprule
    \# & N & Topic & Representation & Example article \\
    \midrule
    1 & 654 & \raggedright LLMs in medical education & medical patient clinical chatgpt questions responses llms health accuracy healthcare potential care education models information ai medicine large performance language study methods gpt4 results bard using data nursing & \citep{Rouhi2024-ku}\\
    2 & 546 & \raggedright Text-to-speech for learning support & speech texttospeech reading students audio text learning technology voice disabilities music people blind visually impaired using software language recognition english comprehension accessibility assistive application study read & \citep{Weerakoon2024-rg}\\
    3 & 249 & \raggedright GenAI in computing education & programming code students llms software models computer introductory tools computing learning course language feedback courses science chatgpt exercises cs1 ai generative problems help education explanations & \citep{Prather2024-uo}\\
    4 & 160 & \raggedright Writing support and feedback & writing students feedback ai generative tools study essays chatgpt english language literacy essay teachers skills research intelligence artificial academic learners llms creative tool texts results business model & \citep{Liu2024-ua}\\
    5 & 139 & \raggedright Ethical and security challenges & ai intelligence artificial generative marketing research ethical chatgpt technologies data potential systems development scientific new human digital technology education risks language article models large llms future security challenges & \citep{Zuber2024-ml}\\
    6 & 129 & \raggedright Educational chatbots & chatgpt chatbots learning students education information ai educational generative user data srl intelligence support artificial knowledge research teachers users college tools teaching academic & \citep{Hedderich2024-mz}\\
    7 & 92 & \raggedright Tools for content and learning design & generative learning ai education educational content genai teachers tools design lesson intelligence personalized teaching artificial students instructional learners study creation materials potential collaborative humanmachine educators classroom experience future create paper & \citep{Powell2024-ot}\\
    8 & 74 & \raggedright Feedback and personalized responses & learning feedback language models llms large knowledge educational model processing systems students natural prompt personalized responses nlp generation peerfeedback llm education instructors engineering content using understanding ai adaptive paper student & \citep{Hutt2024-fy}\\
    9 & 63 & \raggedright Acceptance and adoption & acceptance factors expectancy technology intention students influence use chatgpt study perceived theory ai adoption behavioral significantly usage tools equation satisfaction tam structural effort model unified research generative intentions education genai & \citep{Shahzad2024-gw}\\
    10 & 54 & \raggedright GenAI in higher education & genai higher education learning teaching students gen change generative ai student artt educational chatgpt course experiences adult use research tools university collection study approach technologies impact assessment adoption leadership literacy & \citep{Chiu2024-ju}\\
    11 & 54 & \raggedright Assessment and academic integrity & assessment assessments generative authentic ai learning higher students academic gai tools evaluative integrity student academics education artificial intelligence work exams traditional judgement educational chatgpt formative genai new online practices concerns & \citep{Luo-Jess-2024-vi}\\
    12 & 53 & \raggedright LLMs in education & education learning llms language large models educational ai chatgpt potential intelligence artificial higher challenges generative teaching educators clos ml paper students human article teachers opportunities tools impact particularly tool strategies & \citep{Pelaez-Sanchez2024-qt}\\
    13 & 51 & \raggedright Automated grading and performance evaluation & answer grading models scoring performance model answers assessment item automated large questions student evaluation llms dataset language responses items accuracy gpt4 educational similarity test section short evaluating math formative & \citep{Gao2024-aw}\\
    14 & 49 & \raggedright Ethical implications and policies & ethical ai generative education ethics issues educational intelligence chapter higher use artificial learning policy challenges privacy educators concerns technologies students risks tools emerging implications technology teaching humanities human potential academic & \citep{Nartey2024-fm}\\
    15 & 48 & \raggedright Detecting AI-generated text & detection aigenerated academic text essays texts integrity detectors plagiarism ai writing humanwritten generated articles content detect use student models written tools human work chatgpt accuracy language students distinguishing airephrased educators & \citep{Fleckenstein2024-ts}\\
    16 & 46 & \raggedright Creativity, art, and design & magic art creative ai creativity generative design digital students animation tools images education artistic artificial intelligence divergent thinking study human religious icons media explore texttoimage article create hybrid class new & \citep{Vartiainen2023-yt}\\
    17 & 46 & \raggedright Chatbots for language learning & language chatgpt chatbots learning korean large speaking research teachers chinese chatbot education teaching efl vocabulary based models using oral study interviews students learners potential foreign ai use coole pitanja artificial & \citep{Zhang2024-wo}\\
  \bottomrule
\end{tabular}
\end{table*}

\begin{table*}[htbp!]
  \label{tab:topics2}
  \begin{tabular}{p{0.2cm} p{0.4cm} p{3cm} p{11.3cm} p{0.9cm}}
    \toprule
    \# & N & Topic & Representation & Example article \\
    \midrule
    18 & 43 & \raggedright Sentiments toward AI in social media & social chatgpt resilience chatbot media study emotions educational negative chatbots sentiment childrens analysis education users use coaching trust higher version learning autonomy research positive online intelligence emotional ai findings implications & \citep{Mamo2024-wp}\\
    19 & 43 & \raggedright Impact on academic integrity & integrity academic higher ai plagiarism universities policy genai policies students use institutions ethical education misconduct tools intelligence student artificial institutional generative research heis study staff concerns regarding issues analysis framework & \citep{Lee2024-qv}\\
    20 & 42 & \raggedright AI literacy & ai literacy research generative students study gai higher education academic tools perceptions artificial findings intelligence use participants teachers adoption competence k12 using usage knowledge professionals approach genai ais competencies technology & \citep{Chiu2024-in}\\
    21 & 37 & \raggedright Translation and intercultural skills & translation foreign language teaching issue english special artificial interpreting intelligence papers skills research studies communication competence & \citep{Dai2024-ri}\\
    22 & 36 & \raggedright Automated questions generation & question questions generation mcqs evaluation generated language conference model qg t5 educational quality learning wang kcs prompting models generate aqg large generating automatic knowledge teachers natural using llms qa transformer & \citep{Hang2024-as}\\
    23 & 36 & \raggedright Tutoring systems & tutoring tutor tutors learning dialogue da large students bot models systems student collaborative classifier language instructor llm knowledge chatbot classifiers feedback architecture design support model teachable intelligent dialogues virtual asr & \citep{Lin2024-vf}\\
    24 & 35 & \raggedright Language teaching & genai language teaching tools english study teachers chatgpt students learning qualitative ai generative attitudes use findings educators efl gai education enhance challenges academic ilte integration quantitative analysis interviews research l2 & \citep{Moorhouse2024-hm}\\
    25 & 34 & \raggedright Chemistry education & chemistry chemical engineering students mechanistic reasoning scientific responses problems chatbots valence prompts assignments science chatgpt writing llms molecule simulation generative ai molecular learning artificial & \citep{Tassoti2024-dg}\\
    26 & 34 & \raggedright Text-to-image for visual design and creativity & designers fashion ideation creative process creativity solutions product tools visual thinking texttoimage stimuli images generative ideas analogies collaborative phase search students sustainable study technology engineering & \citep{Jung2024-hg}\\
    27 & 31 & \raggedright Mathematical explanations and problem-solving & mathematical mathematics explanations problems llms models tutoring word feedback language tasks gpt4 large llama answers error model problemsolving learning generated stories lesson generating plans task & \citep{Collins2024-xg}\\
    28 & 29 & \raggedright Teacher education and professional development & teachers gai educators ai teaching tools study training generative practices education mathematics technological preservice participants intelligence learning projectbased artificial research chatgpt support assessment preparedness professional faculty development & \citep{Tan2024-er}\\
    29 & 26 & \raggedright Engineering education & engineering generative ai students tools chatgpt mechanical education circuits peer team teams paper design electric individual learning tool performance civil feedback practices hackathon research career questions study & \citep{Yelamarthi2024-ru}\\
    30 & 25 & \raggedright Tools for library and research & libraries information librarians literacy services academic archival research data faculty entrepreneurs university ai technology students generative media xr gis adoption staff digital tools new legacy justice community & \citep{Deschenes2024-em}\\
    31 & 25 & \raggedright Physics education, problem-solving, and simulations & physics llms language problems large solving models tasks questions model solve problemsolving learning prompting gpt4 students different llm chatgpt instructional generate answers physical simulations research school & \citep{Riabko2024-bi}\\
    32 & 24 & \raggedright Business and entrepreneurial education & business ai entrepreneurship genai generative education intelligence artificial higher future research students case opportunities impact learning potential study entrepreneurial competencies academic service course implications informed teaching & \citep{Aure2024-vu}\\
    33 & 21 & \raggedright Philosophy of education in the age of AI & intelligence human generative ai cognitive artificial connectionist knowledge philosophical education paradox epistemology perils tension biological epistemic argues stiegler philebus technologies embodied cognition nature virtue educational learning & \citep{Class2024-nj}\\
    34 & 20 & \raggedright Geoscience education & visual models image capabilities gis geotechnical problems programming multimodal model gpt4 large spatial reasoning images vision diagram geoscience features exam weather science llms challenges solving earth gpt4s finegrained & \citep{Chen2024-ip}\\
    35 & 20 & \raggedright Critical perspective on chatbots & higher ai chatbots students education critical chatgpt research supervision supervisors practices tools supervisory artificial postgraduate academic chatbot concerns integrity impact generative intelligence need claims ethical teaching faculty study & \citep{Jensen2024-pm}\\
  \bottomrule
\end{tabular}
\end{table*}

\begin{table*}[htbp!]
  \label{tab:topics3}
  \begin{tabular}{p{0.2cm} p{0.4cm} p{3cm} p{11.3cm} p{0.9cm}}
    \toprule
    \# & N & Topic & Representation & Example article \\
    \midrule
    36 & 20 & \raggedright Sentiment and emotion analysis & sentiment emotion student data course annotations classification analysis feedback courses labels models language emotions crowdsourced processing bert text large natural performance comments sentiments stage dataset approach labeled & \citep{Parker2024-wr}\\
    37 & 20 & \raggedright Detecting AI-generated code submissions & code submissions style detectors detection anomaly perplexity anomalies assignments plagiarism cheating models programming detect similarity solutions work large detecting contractors student bounds aigenerated codebert classifier generated tests samples problem constants & \citep{Pan2024-ys}\\
    38 & 17 & \raggedright GenAI and serious games & games escape gamebased rooms plans educational designers framework puzzles design actions players learning generation development levels reactions chemical process procedural content atoms educators & \citep{Tyni2024-wd}\\
  \bottomrule
\end{tabular}
\end{table*}


\begin{acks}

Support from the Research Council of Finland under Grant number 353325 and ANID/ PIA/ Basal Funds for Centers of Excellence FB0003 is gratefully acknowledged.

\end{acks}

\bibliographystyle{ACM-Reference-Format}
\bibliography{bib} 

\end{document}